\documentclass[a4paper,fleqn]{cas-dc}

\usepackage[authoryear]{natbib}
\usepackage{times}
\usepackage{epsfig}
\usepackage{graphicx}
\usepackage{amsmath}
\usepackage{amssymb}
\usepackage{pifont}
\usepackage{multirow}
\usepackage{color}
\usepackage{hyperref}
\usepackage{booktabs}
\usepackage[switch]{lineno}

\def\tsc#1{\csdef{#1}{\textsc{\lowercase{#1}}\xspace}}

\tsc{WGM}
\tsc{QE}

\begin{document}

\let\WriteBookmarks\relax
\def\floatpagepagefraction{1}
\def\textpagefraction{.001}

\shorttitle{Collaborative Enhancement Network for Low-quality Multi-spectral Vehicle Re-identification}    

\shortauthors{\textit{Aihua Zheng ~et al.}}  

\title [mode = title]{Collaborative Enhancement Network for Low-quality Multi-spectral Vehicle Re-identification}

\author[1]{Aihua~Zheng}[orcid = ]
\ead{ahzheng214@foxmail.com}
\credit{Conceptualization of this study and Methodology}

\affiliation[1]{organization={Information Materials and Intelligent Sensing Laboratory of Anhui Province, Anhui Provincial Key Laboratory of Multimodal Cognitive Computation, School of Artificial Intelligence},
            addressline={Anhui University }, 
            city={Hefei},
            postcode={230601}, 
            country={China}}
\affiliation[2]{organization={Anhui Provincial Key Laboratory of Multimodal Cognitive Computation, School of Computer Science and Technology },
addressline={Anhui University}, 
city={Hefei},
postcode={230601}, 
            country={China}}
\affiliation[3]{organization={School of Biomedical Engineering},
            addressline={Anhui Medical University}, 
            city={Hefei},
            postcode={230032}, 
            country={China}}
\author[1]{Yongqi~Sun}[orcid =]

\ead{yong-qi-sun@foxmail.com}

\credit{Investigation, Writing-Original Draft, Validation and Visualization}

\author[3]{Zi~Wang}[orcid =]
\ead{ziwang1121@foxmail.com}
\credit{Writing Review and Editing}

\author[1]{Chenglong~Li*}[orcid = 0000-0002-7233-2739]
\ead{lcl1314@foxmail.com}
\credit{Formal Analysis and Data Curation  }

\author[2]{Jin~Tang}[orcid =]
\ead{tangjin@ahu.edu.cn}
\credit{Resources and interpretation of data}

\cortext[1]{Corresponding Author: Chenglong Li.}

\begin{highlights}

\item A new collaborative enhancement framework for low-quality multi-spectral vehicle re-identification.
		
\item  A proxy generator to aggregate residual key information and generate fused proxy.
				
\item A dynamic quality sort module to dynamically identify high-quality primary spectra based on information relevance.

\item A collaborative enhancement module complements spectra with missing discriminative cues and detail information.

\end{highlights}


\maketitle

\begin{abstract}
The performance of multi-spectral vehicle Re-identification (ReID) is significantly degraded when some important discriminative cues in visible, near infrared and thermal infrared spectra are lost.
Existing methods generate or enhance missing details in low-quality spectra data using the high-quality one, generally called the primary spectrum, but how to justify the primary spectrum is a challenging problem. 
In addition, when the quality of the primary spectrum is low, the enhancement effect would be greatly degraded, thus limiting the performance of multi-spectral vehicle ReID.
To address these problems, we propose the Collaborative Enhancement Network (CoEN), which generates a high-quality proxy from all spectra data and leverages it to supervise the selection of primary spectrum and enhance all spectra features in a collaborative manner, for robust multi-spectral vehicle ReID.
First, to integrate the rich cues from all spectra data, we design the Proxy Generator (PG) to progressively aggregate multi-spectral features.
Second, we design the Dynamic Quality Sort Module (DQSM), which sorts all spectra data by measuring their correlations with the proxy, to accurately select the primary spectra with the highest correlation.
Finally, we design the Collaborative Enhancement Module (CEM) to effectively compensate for missing contents of all spectra by collaborating the primary spectra and the proxy, thereby mitigating the impact of low-quality primary spectra.
Extensive experiments on three benchmark datasets are conducted to validate the efficacy of the proposed approach against other multi-spectral vehicle ReID methods.
{The codes will be released at \href{https://github.com/yongqisun/CoEN}{https://github.com/yongqisun/CoEN.}}
\end{abstract}

\section{Introduction}\label{}
Traditional vehicle ReID tasks usually use only single visible spectral images for retrieval.
Although most single-spectral vehicle ReID methods~\cite{liu2016deepID,guo2018learning,Lou2019veriwild,Tang2019cityflow,shen2023triplet} have demonstrated excellent performance, in complex scenes like nighttime, fog, and flare, the lack of key discriminatory information limits the recognition of single RGB spectra.
In recent years, vehicle ReID tasks based on multi-spectral~\cite{li2020multi} images have emerged.
Visible (RGB) spectra provide rich color information, near-infrared (NIR) spectra are suitable for nighttime or low-light environments, and thermal-infrared (TIR) spectra ignore the effect of light and reflect the thermal radiation state of the vehicle.
Multiple spectra provide diverse vehicle details and complementary advantages, and studies~\cite{zheng2023cross,zheng2023visible,huangdeep2023,ZHENGmulti2024,HUANG2024102194,wang2025DeMo,wang2025MambaPro} have been conducted to achieve excellent performance by reducing domain differences between spectra and using multi-spectral information for complementary enhancement.
\begin{figure}
\centering
\includegraphics[width=1.0\linewidth]{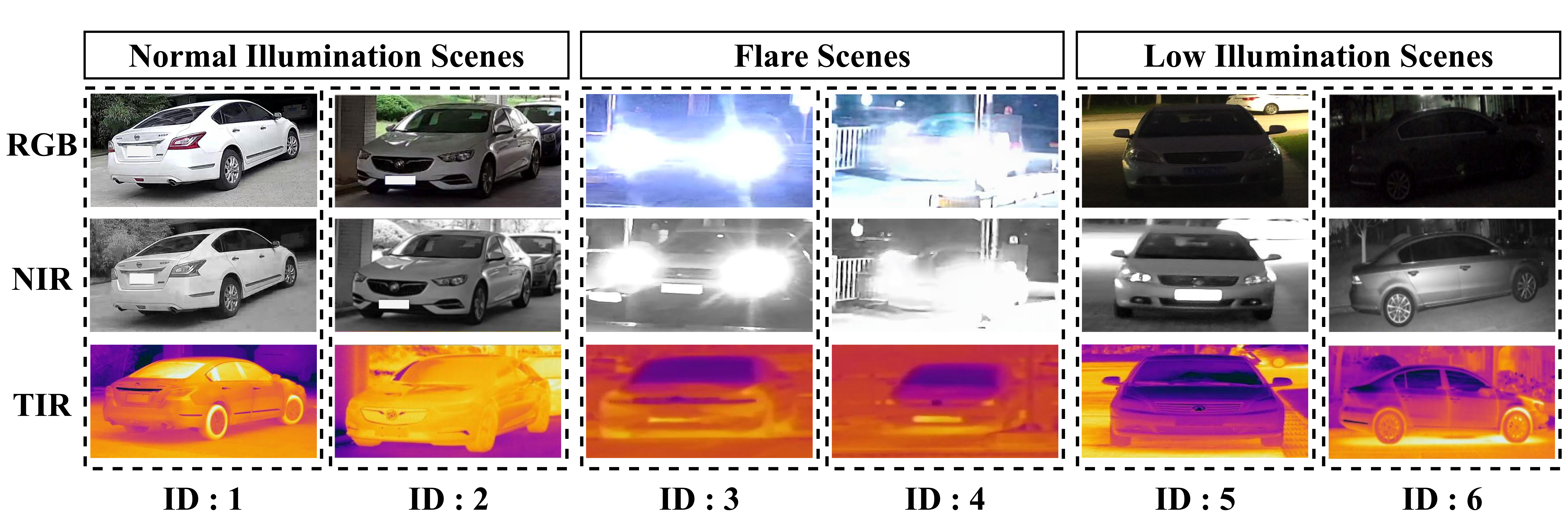}
  \vspace{-5mm}
  \caption{Performance of different spectra in complex scenes. RGB and NIR spectra perform better in normal illumination scenes, TIR spectra contain more identity information in flare scenes, while NIR and TIR spectra show superior performance in low illumination scenes}
  \label{fig:motivation1}
  \vspace{-4mm}
\end{figure}

However, different spectra inevitably lead to the loss of important discriminative cues~\cite{facenet,wang2025DeMo} for vehicles due to their specific imaging characteristics, which impairs the performance of multi-spectral vehicle ReID.
Especially in complex scenes, the quality varies significantly between spectra.
For example, as shown in Fig.~\ref{fig:motivation1}, the RGB and NIR spectrum outperforms the TIR with normal illumination due to their rich color information. By contrast,  the image quality of the RGB and NIR spectra deteriorates significantly under strong flares~\cite{facenet}, while the TIR spectrum remains relatively high in quality since it is unaffected by the flares. Under low illumination scenarios, color and texture information is severely missing from the RGB spectrum, with NIR and TIR showing more robust quality.

\begin{figure}[t]
\centering
\includegraphics[width=0.95\linewidth]{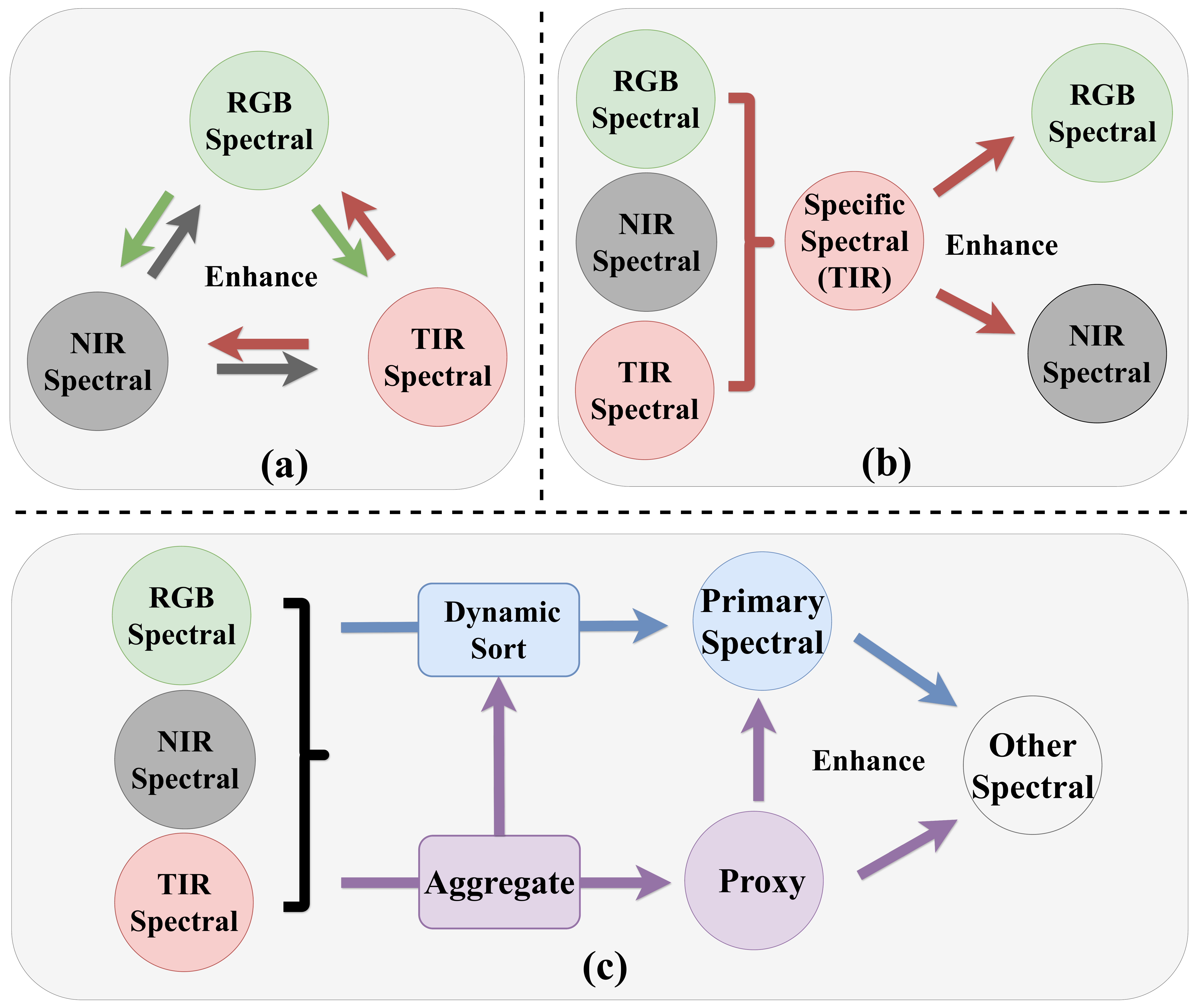}
  \vspace{-1mm}
  \caption{(a) Mutual-based enhancement method~\cite{topreid,wang2025DeMo,wang2025MambaPro}. (b) Specific spectral-based primary enhancement method~\cite{facenet}. (c) Our method.}
  \label{fig:motivation2}
  \vspace{-4mm}
\end{figure}

These quality differences exacerbate domain discrepancies between spectra, posing significant challenges for multi-spectral vehicle ReID.
In real-world scenarios, low-quality spectra often contain more noise, leading to more ambiguous and unstable vehicle features.
As shown in Fig.~\ref{fig:motivation2} (a), traditional mutual-based enhancement solutions~\cite{topreid,wang2025DeMo,wang2025MambaPro} generate or enhance spectra of varying quality through interactions between all spectra.
However, these approaches overlook noise interference in low-quality spectra on high-quality spectral features, making it unsuitable for scenarios with significant spectral quality differences.
To address these challenges and differentiate from traditional methods, existing studies focus on enhancing discriminative power by compensating for missing information in low-quality multi-spectral data.
As shown in Fig.~\ref{fig:motivation2} (b), Specific spectral-based primary enhancement method~\cite{facenet} selects high-quality TIR spectra as primary spectra and enhances the missing vehicle discrimination information in low-quality spectra.

Although high-quality spectra usually contain more vehicle detail cues, in challenging scenarios where all spectra are of lower quality, relying solely on high-quality spectral enhancement becomes significantly less effective, even failing to provide critical discriminative information.
To this end, we propose an efficient approach that fuses key cues from all spectra to generate fused features enriched with multi-spectral information, thereby enhancing the remaining spectra.
However, this multi-spectral fusion lacks supervision. Although it integrates the remaining discriminative cues, it suppresses the representation of vehicle detail information.
This causes fusion enhancement to overemphasize detail-independent vehicle cues, suppressing the original detail information in the low-quality spectra.
Therefore, preserving the detailed vehicle information while introducing richer discriminative cues to the lower quality spectra is equally crucial.

To address the above issues, as shown in Fig.~\ref{fig:motivation2} (c), our method integrates multi-spectral information to generate proxy, dynamically selects the primary spectra with higher accuracy, and collaborates on two enhancement strategies to complement missing discriminative cues and reduce quality differences between spectra.
Specifically, our CoEN comprises three key components: the Proxy Generator (PG), the Dynamic Quality Sort Module (DQSM), and the Collaborative Enhancement Module (CEM).
First, to effectively integrate the remaining information across all spectra, we design the PG module. This module aggregates discriminative information using a progressive fusion strategy to generate a unified proxy feature. Subsequently, the proxy will be used for subsequent primary spectrum selection and identity inference.
Secondly, each spectrum exhibits significant quality differences due to varying degrees of missing key cues. To accurately assess the quality of each spectrum and select the primary spectrum, we propose the DQSM module. This module dynamically determines the remaining identity information content by measuring the cue correlation between the proxy and each spectrum. Then it generates a quality score, ranks the spectra, assigns quality labels, and ultimately selects the primary spectrum with the highest score.

Thirdly, low-quality spectra suffer from a severe lack of critical cues, which exacerbates the inherent domain differences between spectra. To effectively supplement these cues and mitigate the effects of low-quality primary spectra, we design the CEM to alleviate the discrepancies between different quality spectra and align multi-spectral features. This module complements and enhances vehicle cues in low-quality spectra by introducing rich discriminative cues from proxy, leveraging them alongside detailed information from primary spectra, and preventing the suppression of such details.
CEM employs two enhancement strategies. First, to transfer unique detail information from high-quality spectra to low-quality ones, we design the primary-based enhancement strategy. This strategy calculates the contextual relationship between the primary spectra selected by DQSM and the remaining low-quality spectra, selecting high-confidence discriminative features to migrate to the low-quality spectra. However, when all spectra are of low quality, the effectiveness of this strategy is significantly diminished because the primary spectrum also lacks critical information.
We propose a proxy-based enhancement strategy to reduce information loss across all spectra by utilizing proxy enriched with multi-spectral information to address this issue. Since the generated fusion features might overlook certain detailed information, we integrate the two enhancement strategies to ensure the acquisition of semantically rich multi-spectral feature representations.

In summary, our contributions are as follows:
\begin{itemize}
  \item
  We propose a new Collaborative Enhancement Network CoEN for multi-spectral vehicle ReID.
  To the best of our knowledge, this is the first work that considers combining features for low-quality multi-spectral vehicle ReID.
  \item
  We propose a Proxy Generator (PG) that creates a shared proxy metric for multi-spectra via progressive aggregation. Additionally, we propose the Dynamic Quality Sort Module (DQSM), which assigns quality labels to spectra based on calculated correlation scores.
  \item
  We propose the Collaborative Enhancement Module (CEM), which collaborates two enhancement strategies to improve low-quality spectra, complement missing discriminative cues and detail information, and align multi-spectral features.
  \item
  We conduct comprehensive experiments on three multi-spectral vehicle ReID benchmarks WMVeID863 ~\cite{facenet}, RGBNT100~\cite{li2020multi}, and MSVR310~\cite{zheng2023cross}.
  The results fully validate the effectiveness of our proposed approach.
\end{itemize}

\section{Related Work}
\subsection{Single-spectral Vehicle ReID}
In recent years, with the development of ReID tasks, single-spectral vehicle ReID has received increasing attention.
Existing single-spectral vehicle ReID methods are mainly dominated by CNN-based~\cite{he2016deep} methods~\cite{liu2016deepID,meng2020parsing,chen2020orientation,zhao2021phd,chen2022sjdl,li2022vehicle,li2022attribute,shen2023git,li2024day}.
~\cite{liu2016deepID,liu2016deep} first propose two large-scale vehicle datasets Veri776 and VehicleID with multiple vehicle identities and spatio-temporal labels.
~\cite{meng2020parsing} propose a view-aware embedding network based on parsing (PVEN) to address the problem of view-aware feature alignment and enhancement. By separating vehicles into multiple views and aligning features, the intra-instance distance is reduced and the inter-instance differences are increased through a shared visible attention mechanism.
~\cite{chen2020orientation} propose a semantics-guided part attention network (SPAN) to generate attention masks for vehicle parts with different viewpoints, extracting the discriminative features of each part separately.
~\cite{li2022attribute} propose an attribute and state guided structural embedding network (ASSEN) to improve vehicle ReID feature learning by amplifying identity-related vehicle attributes to increase inter-class differences and reduce intra-class variations by suppressing the impact of state-related interference.
~\cite{li2024day} introduce a new cross-domain dataset, DNWild, and design a novel day-night dual-domain modulation (DNDM) framework that reduces vehicle glare at night while enhancing vehicle features in low-light conditions, thereby alleviating the domain gap between day and night images.

However, CNN~\cite{he2016deep} struggles to capture global information effectively due to their limited receptive field, whereas ViT~\cite{dosovitskiy2020image} is gaining popularity for its strong global perception capabilities.
~\cite{he2021transreid} encode images as patch sequences and construct a strong Transformer-based baseline model. It enhances feature discriminability and coverage through patch shifting and shuffling while introducing learnable viewpoint embeddings to mitigate intra-class bias caused by viewpoint differences.
Subsequently, Transformer-based methods prove the effectiveness of the self-attention mechanism on the re-recognition method.
~\cite{shen2023git} proposes the graph interactive transformer (GIT), which enhances vehicle ReID performance by leveraging a graph to extract discriminative local features from patches and a transformer to capture robust global features.
However, these methods do not consider the effects of all-weather complex scenes and only process single spectra with limited representational power.
As a result, they can only represent limited discriminative information.

\subsection{Multi-spectral Object ReID}
Multi-spectral object ReID leverages data from different spectra, including RGB, NIR, and TIR, to improve feature fusion and enhance ReID task performance.
Firstly, ~\cite{li2020multi} propose two multi-spectral vehicle datasets, RGBN300 and RGBNT100, and construct a baseline method HAMNet, which facilitates multi-spectral information fusion using CAM~\cite{zhou2016learning}.
~\cite{zheng2023cross} propose a cross-directional consistency network (CCNet) and introduce a cross-directional center loss to address spectral and sample differences, overcoming the challenges of vehicle ReID in complex lighting and diverse scenarios, and present a new multi-spectral vehicle ReID benchmark, MSVR310.
To leverage the complementary strengths of multi-spectral data, ~\cite{zheng2021robust} design a new progressive fusion network that reduces modal differences by using a single spectrum to learn effective multi-spectral representations from multiple spectra.
~\cite{guo2022generative} reduce spectral differences by generating transition spectra.
~\cite{wang2022interact} further consider the importance of retaining specific spectral information for feature fusion, highlighting this by expanding inter-spectral differences after fusing information from different spectra.
~\cite{he2023graph} employ a graph convolutional network within an end-to-end learning framework to fuse multi-modality features adaptively.
~\cite{Kamenou2023CVPR} address the challenge of limited infrared spectra data availability for multi-spectral vehicle ReID by proposing a domain generalization approach that leverages meta-learning.
~\cite{pang2024inter} encourage the learning of modal invariance and identity features by emphasizing inter-modal similarity and shape information.

For Transformer-based methods,
~\cite{topreid} propose a recursive marker substitution framework to make full use of the local details of different spectra to generate more discriminative multi-spectral features.
~\cite{magictoken} adaptively select object-centered markers based on the spatial and frequency information of the spectrum.
However, these methods generate and supplement information primarily by treating all spectra as primary spectra.
~\cite{facenet} introduce the first ReID dataset for low-quality multi-spectral vehicles affected by flares, using the thermal infrared spectrum, which remains robust under flare conditions, as the primary spectrum to complement and enhance discriminative information in other spectra.
However, this approach selects the primary spectrum based on data prior, which may not be suitable for other complex environments, and it does not account for the impact of full-spectrum low quality.
%

\section{Collaborative Enhancement Network}
\begin{figure*}
    
\centering
\begin{center}
  \includegraphics[width=1.95\columnwidth]{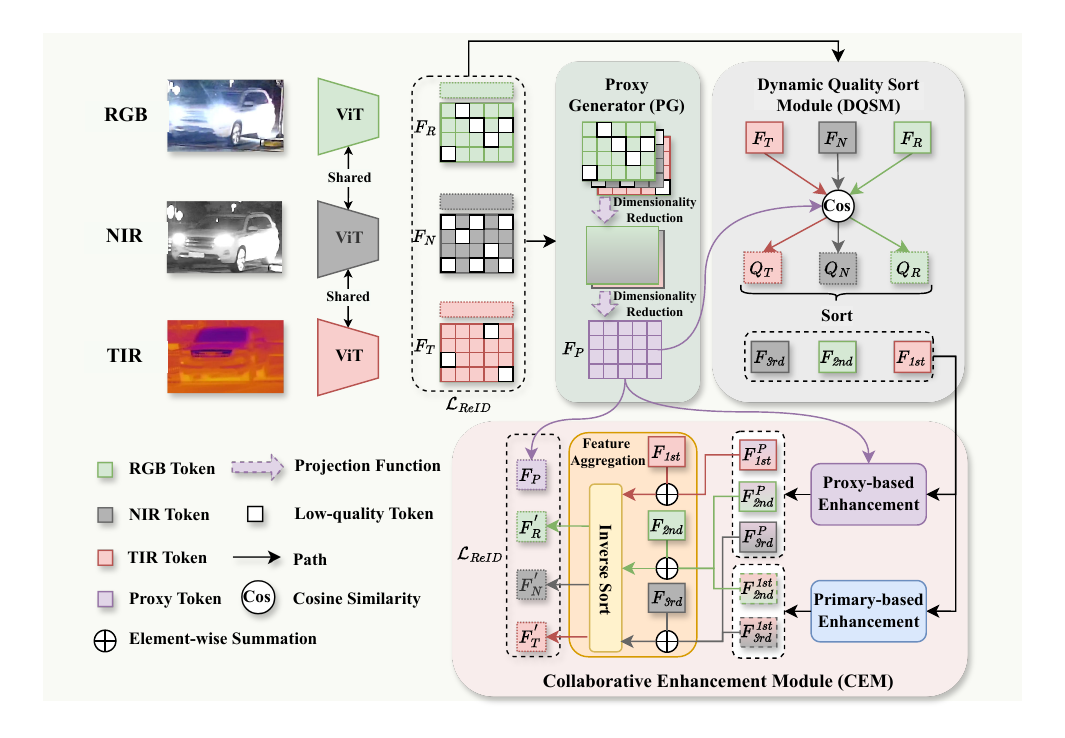}
  \end{center}
  \vspace{-10mm}
  \caption{
  The framework of our proposed Collaborative Enhancement Network (CoEN). First, features ($F_R$, $F_N$, $F_T$) from different spectra are extracted using a ViT~\cite{dosovitskiy2020image} backbone network with shared parameters. These features are fed into the Proxy Generator (PG) to generate fused proxy feature $F_P$. Next, the Dynamic Quality Sort Module (DQSM) sorts and selects high-quality spectra feature $F_{\textit{1st}}$. Finally, the Collaborative Enhancement Module (CEM) enhances low-quality spectra by complementing missing discriminative cues and detail information, yielding multi-spectral features with rich discriminative information for vehicle ReID.
  }
  \label{fig:Overall}
  \vspace{-2mm}
\end{figure*}

To collaboratively leverage vehicle details in primary spectra and rich discriminative signals in combined spectral features, while compensating for missing signals in low-quality spectra and reducing domain discrepancies caused by varying qualities, we propose the Collaborative Enhancement Network (CoEN).
As shown in the Fig.~\ref{fig:Overall}, our proposed CoEN consists of four key components: Shared Feature Extraction, Proxy Generator (PG), Dynamic Quality Sort Module (DQSM), and Collaborative Enhancement Module (CEM).
We will describe four key modules in the following subsections.
\subsection{Shared Feature Extraction}
To efficiently extract multi-spectral features while addressing the discrepancies between different spectra and minimizing the model parameters, we design and implement a shared Vision Transformer (ViT)~\cite{dosovitskiy2020image} to process the multi-spectral inputs. For each spectral (including RGB, NIR, and TIR), we extract the corresponding multi-spectral features individually, 
\begin{equation} 
    F_{R}, F_{N}, F_{T} = \mathrm{ViT}\left(S_{R}, S_{N}, S_{T}\right),
\end{equation}
\noindent where $S_R$, $S_N$, and $S_T$ represent the RGB, NIR, and TIR input spectral images, respectively.
The tokenized features $F_R$, $F_N$, and $F_T$, each of which has a shape of $\mathbb{R}^{(N_{p}+1)} \times D$, are extracted from the penultimate layer of $\mathrm{ViT}$, we keep the last layer of blocks to integrate the enhanced spectral features.
$N_{p}$ means the number of patch tokens while $D$ is the embedding dimension.
Specifically, to obtain more informative feature representations, we select the embedding tokens from the backbone network for combined feature generation and feature enhancement. Moreover, we employ a ViT Block~\cite{dosovitskiy2020image} layer at the end to extract rich discriminative information from the enhanced multi-spectral embedding tokens and translate it into global class tokens.

\subsection{Proxy Generator}
The loss of discriminative information varies across different environments, resulting in significant differences in the quality of the spectra. 
However, each spectrum still retains some residual vehicle information.
Due to the limited vehicle information in a single spectrum, we propose the Proxy Generator (PG) to fully utilize the complementary advantages of multiple spectra.
This module combines different spectra while minimizing information loss by applying a progressive feature aggregation strategy to generate a unified proxy representation.
Additionally, the fused proxy features contain rich vehicle discriminative information and play a key role in the proxy-based enhancement strategy and the final inference stage.
Specifically, we adopt a gradual dimensionality reduction strategy to map the multi-spectral information into a unified feature space, effectively integrating the information from each spectrum.
As shown in Fig.~\ref{fig:Overall}, we first choose embedding tokens $F_R^{emb}$, $F_N^{emb}$ and $F_T^{emb} \in \mathbf{R}^{D \times {N_{p}}}$ from $F_{R}$, $F_{N}$ and $F_{T}$. Since the embedding tokens capture finer-grained global spectral representations, we concatenate the multi-spectral embedding tokens along their dimensions to ensure fusion quality and generate the initial high-dimensional proxy,
\begin{equation}
    F^{Init}_{P} = Concat(F_{R}^{emb},F_{N}^{emb},F_{T}^{emb}), F^{Init}_{P}\in\mathbb{R}^{3D\times{N_{p}}},
\end{equation}
where $F_{i}^{emb}$ stands for the embedding tokens for spectral $i$, $i\in\{R, N, T\}$. $Concat$ stands for connection operations in PyTorch.
At this stage, the initial proxy features embed multiple spectral information into the high-dimensional feature space, but the remaining cues from each spectrum remain dispersed across different spectra.
Therefore, we remap the obtained $F^{Init}_{P}$ to the single-spectrum feature space to integrate the remaining spectral cues.
To prevent semantic loss caused by abrupt dimensionality reduction, we employ a progressive dimensionality reduction strategy to mitigate high-dimensional semantic collapse while preserving discriminative information in the fused features.
Specifically, we apply a two-step reduction in dimensionality to the initial proxy features to optimize information aggregation and feature mapping,
\begin{equation}
    F_{P} = Proj_B(Proj_A(F^{Init}_{P})), F_{P}\in\mathbb{R}^{D\times{N_{p}}},
\end{equation}
where $F_{P}$ represents the proxy features for a single sample, $Proj_A$ and $Proj_B$ represents the projection layers.
Through progressive dimensionality transformation, we finally aggregate the discrete multi-spectral residual key features into proxy features $F_{P}$ that contain rich multi-spectral information.
$F_{P}$ contains the identity cues in different spectra.
Therefore, $F_{P}$ can be used for spectral quality assessment and convey their rich semantic information to individual spectra to enhance feature representation.

\subsection{Dynamic Quality Sort Module}
Although FACENet~\cite{facenet} relies on visual priors to classify TIR spectra as high quality, this approach is only effective in flare environments and struggles with other complex scenarios.
With the lack of quality labels, accurately selecting the highest-quality spectra remains a major challenge.
To accurately select the highest-quality spectra across different scenes, we design the Dynamic Quality Sort Module (DQSM). This module dynamically evaluates the quality of each spectrum as it varies with the scene, using the proxy generated by PG as a benchmark. Specifically, DQSM computes the correlation between the proxy feature vector and each spectral feature vector, assigns a correlation score to each spectrum, and ranks them to select the high-quality spectra containing the most remaining vehicle cues.
Taking RGB spectral features as an example, to accurately measure their relevance to the fusion features, we sequentially extract token vector pairs from the $N_P$ tokens of both and compute their cosine similarity in the feature space. For the $i_{th}$ vector pair, 
\begin{equation}
    S_i(a, b) = \frac{\sum_{j=1}^{D} a_{i,j} \cdot b_{i,j}}{\sqrt{\sum_{j=1}^{D} a_{i,j}^2} \cdot \sqrt{\sum_{j=1}^{D} b_{i,j}^2}},
\end{equation}
where $a$ and $b$ represent features $F_P$ and $F_{R}^{emb}$ respectively, $S_{i}(a, b)$ represents the cosine similarity of the $i$-th token vector pair, $D$ represents the feature dimension. From this, we obtain a similarity vector $S$ of dimension $N_P$. 
Then, we compute the average of all elements in this similarity vector $S$ to obtain the average relevance score of all token pairs. This value is used as a measure of overall feature similarity, and thus, the quality scores of each spectral feature are derived, 
\begin{equation}
    Q_R = \bar{S}(F_P, F_{R}^{emb}) = \frac{1}{N_{P}} \sum_{i=1}^{N_{P}} S_i(F_P, F_R^{emb}),
\end{equation}
where $Q_R$ represents the quality scores of RGB spectra, $\bar{S}$ represents the average operation on the vector $S$. By performing fine-grained information correlation analysis, we accurately compute the quality score of the RGB spectrum and apply the same method to calculate the quality scores of the NIR and TIR spectra, 
\begin{equation}
    Q_N = \bar{S}(F_P, F_{N}^{emb}), Q_T = \bar{S}(F_P, F_{T}^{emb}).
\end{equation}

Finally, we rank the quality scores of each spectrum to obtain a spectrum quality ranking, where a larger quality score means that the spectrum contains richer discriminative information,
\begin{equation}
    F_{\textit{1st}}, F_{\textit{2nd}}, F_{\textit{3rd}} = Sort(F_R^{emb}, F_N^{emb}, F_T^{emb}, Q_R, Q_N, Q_T),
\end{equation}
where $F_{\textit{1st}}$ represents the spectrum feature with the highest quality score, and so on.
Our CoEN will adaptively enhance each spectrum based on the ranking information.

\subsection{Collaborative Enhancement Module}
To complement and enhance missing vehicle critical cues in the spectra while mitigating domain discrepancies among multiple spectra, we introduce the Collaborative Enhancement Module (CEM). 
This module leverages proxy and primary spectral to compensate for missing information in each spectrum by extracting rich discriminative cues from proxy and detailed information from high-quality primary spectra.
Specifically, CEM comprises two components: primary-based enhancement and proxy-based enhancement. 
Primary-based enhancement utilizes the primary spectra selected by DQSM to compensate for missing detailed information in low-quality spectra, relying on robust spectral features extracted from real images. 
Proxy-based enhancement, on the other hand, leverages proxy generated by PG, which contains rich discriminative information to compensate for missing information in each spectrum. 
However, the effectiveness of primary-based enhancement is constrained when the primary spectra lack sufficient information, while proxy-based enhancement is limited by the absence of certain fine-grained details in the fused proxy.
Therefore, CEM integrates both enhancement strategies to fully exploit their respective strengths, improving overall spectral quality and enhancing ReID performance.
\\
\textbf{Primary-based Enhancement.}
\begin{figure}[t]
  \centering
  \includegraphics[width=0.95\linewidth]{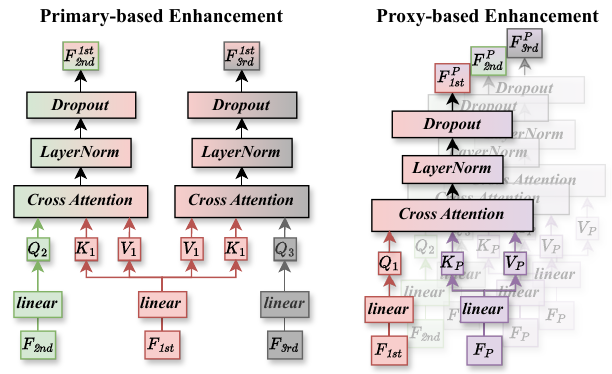}
  \vspace{-1mm}
  \caption{Illustration of Collaborative Enhancement Module.}
  \label{fig:cem}
  \vspace{-4mm}
\end{figure}
As illustrated in Fig.~\ref{fig:cem}, primary-based enhancement extracts identity-related information from high-quality spectra to improve and complement low-quality spectra. 
Specifically, it captures high-confidence details by aligning the contextual information of the low-quality spectrum with that of the primary spectrum.
For $F_{\textit{1st}}$ and $F_{\textit{2nd}}$, we use the low-quality spectra $F_{\textit{2nd}}$ as query $Q_2$ and the primary spectra $F_{\textit{1st}}$ as key $K_1$ and value $V_1$,
\begin{equation}
    Q_2 = F_{\textit{2nd}} \odot W, K_1 = F_{\textit{1st}} \odot W, V_1 = F_{\textit{1st}} \odot W,
\end{equation}
where $Q_2, K_1, V_1 \in\mathbb{R}^{N_{p}\times D}$, and $W \in\mathbb{R}^{D\times D}$ represents the weight matrix. 
Next, we calculate the similarity between the query matrix $Q_2$ and the key matrix $K_1$. Then, we apply the softmax function to normalize it, focusing on regions with a high correlation between the main and current spectra. The resulting matrix is multiplied by the value matrix $V_1$ to extract fine-grained features from the high-quality spectrum. To mitigate noise and prevent overfitting of the model, we drop some data, ultimately obtaining the main enhanced features $F_{\textit{2nd}}^{\textit{1st}}$, 
\begin{equation}
    F_{\textit{2nd}}^{\textit{1st}} = \mathrm{Dropout_\gamma} \left(Softmax \left(\frac{Q_2K_1^T}{\sqrt{d_C}}\right)V_1\right),
    \label{eq:pe}
\end{equation}
where $F_{\textit{2nd}}^{\textit{1st}}$ denotes the enhanced feature obtained after augmenting $F_{\textit{2nd}}$ with $F_{\textit{1st}}$, $\mathrm{Dropout_\gamma}(\cdot)$ denotes dropout operation, $\gamma$ denotes dropout rate. Here, $\gamma=0.5$ and $d_C=D$.
Similarly, we can get the enhancement feature $F_{\textit{3rd}}^{\textit{1st}}$ for low-quality spectra $F_{\textit{3rd}}$.
\\
\textbf{Proxy-based Enhancement.}
When the primary spectra also suffer from significant missing key information, the primary-based enhancement strategy alone cannot effectively compensate for the missing information in low-quality spectra.
Even the primary spectra themselves require enhancement.
Therefore, we design the proxy-based enhancement strategy to supplement key features for all spectra using the full spectral representation $F_P$ generated by the PG.
Specifically, as shown in Fig.~\ref{fig:cem} for the spectrum $\textit{1st}$, we use it as the query matrix $Q_1$ and the proxy features as the key matrix $K_P$ and the value matrix $V_P$, in the case of $F_{\textit{1st}}$, we obtain the enhancement feature $F_{\textit{1st}}^{P}$ for $F_{\textit{1st}}$,

\begin{equation}
    F_{\textit{1st}}^{P} = \mathrm{Dropout_\gamma} \left(Softmax \left(\frac{Q_1K_P^T}{\sqrt{d_C}}\right)V_P\right).
    \label{eq:fe}
\end{equation}

Similarly, we can get the enhancement feature $F_{\textit{2nd}}^{P}$ and $F_{\textit{3rd}}^{P}$ for spectra $F_{\textit{2nd}}$ and $F_{\textit{3rd}}$.
We extract rich vehicle discriminative cues, by aligning the proxy features with individual spectral features.
\\
\textbf{Feature Aggregation.}
As shown in Fig.~\ref{fig:Overall}, we aggregate the enhanced features obtained from primary-based enhancement and proxy-based enhancement. To ensure that the enhanced spectral features retain the spectrum-specific information, we introduce the residual join operation. From this, we perform matrix addition on the features of different spectra to obtain the final spectrally enhanced features $F_{\textit{1st}}^{'}, F_{\textit{2nd}}^{'}, F_{\textit{3rd}}^{'}$,
\begin{equation}
    \begin{split}
        F_{\textit{1st}}^{'} &= F_{\textit{1st}} + F_{\textit{1st}}^{P}, 
        \\
        F_{\textit{2nd}}^{'} &= F_{\textit{2nd}} + F_{\textit{2nd}}^{P} + F_{\textit{2nd}}^{\textit{1st}}, 
        \\
        F_{\textit{3rd}}^{'} &= F_{\textit{3rd}} + F_{\textit{3rd}}^{P} + F_{\textit{3rd}}^{\textit{1st}}.
    \end{split}
\end{equation}

\begin{table*}[!ht]
\caption{Comparison with state-of-the-art methods for multi-spectral ReID on WMVeID863 (in \%). Both single-spectral and multi-spectral methods are included.}
\centering
\resizebox{0.8\linewidth}{!}{
\begin{tabular}{cccccccc}
\hline
\multirow{2}{*}{} & \multirow{2}{*}{Methods} & \multirow{2}{*}{Ref} & \multirow{2}{*}{Bac} & \multicolumn{4}{c}{WMVeID863} \\ \cline{5-8} 
    &        &       &      & mAP & R-1 & R-5 & R-10 \\ \hline
    \multirow{8}{*}{Single}  
    & DenseNet~\cite{huang2017densely}      &CVPR'17  & CNN   & 42.9 & 47.9 & 61.9 & 68.7   \\      
    & ShuffleNet~\cite{zhang2018shufflenet} &CVPR'18  & CNN   & 34.2 & 37.2 & 52.3 & 58.9   \\ 
    & MLFN~\cite{chang2018multi}            &CVPR'18  & CNN   & 43.7 & 47.0 & 61.7 & 69.8   \\ 
    & HACNN~\cite{li2018harmonious}         &CVPR'18  & CNN   & 46.9 & 48.9 & 66.9 & 73.8   \\ 
    & BoT~\cite{luo2019bag}                 &CVPR'19  & CNN   & 51.1 & 55.7 & 69.8 & 74.7   \\
    & OSNet~\cite{zhou2019omni}             &ICCV'19  & CNN   & 42.9 & 46.8 & 61.9 & 69.4   \\ 
    & ViT~ \cite{dosovitskiy2020image}      &Arxiv'20 & ViT   & 66.5 & 73.7 & 80.6 & 82.6   \\
    & AGW~\cite{ye2021deep}                 &TPAMI'21 & CNN   & 30.3 & 35.3 & 43.3 & 46.5   \\
    & TransReID~\cite{he2021transreid}      &ICCV'21  & ViT   & 67.0 & 74.7 & 79.5 & 82.4   \\
    & PFD~\cite{wang2022pose}               &AAAI'22  & ViT   & 50.2 & 55.3 & 69.8 & 75.3   \\ \hline
\multirow{9}{*}{Multiple}  
    & HAMNet~\cite{li2020multi}             &AAAI'20  & CNN   & 45.6 & 48.5 & 63.1 & 68.8  \\
    & PFNet~\cite{zheng2021robust}          &AAAI'21  & CNN   & 50.1 & 55.9 & 68.7 & 75.1  \\
    & IEEE~\cite{wang2022interact}          &AAAI'22  & CNN   & 45.9 & 48.6 & 64.3 & 67.9  \\
    & CCNet~\cite{zheng2023cross}           &INFFUS'23& CNN   & 50.3 & 52.7 & 69.6 & 75.1  \\
    & TOP-ReID~\cite{topreid}               &AAAI'24  & ViT   & 67.7 & 75.3 & 80.8 & 83.5  \\
    & EDITOR~\cite{magictoken}              &CVPR'24  & ViT   & 65.6 & 73.8 & 80.0 & 82.3   \\
    & FACENet~\cite{facenet}                &INFFUS'25& ViT   & \underline{69.8} & \underline{77.0} & \underline{81.0} & \underline{84.2}  \\
    & DeMo~\cite{wang2025DeMo}              &AAAI'25  & ViT   & 65.5 & 71.9 & 78.8 & 82.4  \\
    & \bf {CoEN}    & Ours & ViT           &\bf{71.4}& \bf{79.2}& \bf{83.5}& \bf{85.6}  \\ \hline
    
\multirow{3}{*}{Multiple}  
    & DeMo~\cite{wang2025DeMo}              &AAAI'25  & CLIP  & 68.8 & \underline{77.2} & \underline{81.5} & 83.5  \\
    & MambaPro~\cite{wang2025MambaPro}      &AAAI'25  & CLIP  & \underline{69.5} & 76.9 & 80.6 & \underline{83.8}  \\
    & \bf {CoEN}                            & Ours    & CLIP  &\bf{70.9}& \bf{77.9}& \bf{82.7}& \bf{85.6}  \\
    \hline
\end{tabular}
}
\label{Tab:Sotaof863}
\end{table*}

To facilitate the differentiation, we map the obtained spectrally enhanced features back to the spectra according to the sorting information in the DQSM,
\begin{equation}
    F_{R}^{emb'}, F_{N}^{emb'}, F_{T}^{emb'} = IS(F_{\textit{1st}}^{'}, F_{\textit{2nd}}^{'}, F_{\textit{3nd}}^{'}, Q_R, Q_N, Q_T),
\end{equation}
where $IS(\cdot)$ stands for inverse sort operation.
Finally, we use mutually independent single-layer ViT Block~\cite{dosovitskiy2020image} to capture the global information in each spectral embedding tokens $F_{R}^{emb'}, F_{N}^{emb'}$ and $F_{T}^{emb'}$, which contain rich semantic information after enhancement, into class tokens $F_{R}^{cls'}, F_{N}^{cls'}$ and $F_{T}^{cls'}$ for the final ReID task.

\subsection{Overall Loss Function}
In the training stage, the backbone features, the CEM output features, and the proxy features are under the joint supervision of the identity loss~\cite{szegedy2016rethinking} and the triplet loss~\cite{triplet}.

The identity loss~\cite{szegedy2016rethinking} is calculated as the cross entropy between the predicting probability and is denoted as $\mathcal{L}_{id}$. 

\begin{equation}
  \mathcal{L}_{id}^{}(y) = - {\sum_{i=1}^{N}}\sum_{j=1}^{C} \hat{y}^{n}\log_{}{p({y}_{j}^{n})},
  \label{eq:celoss}
\end{equation}
where $\hat{y}^{n}$ is a one-hot matrix indicates the identity label of the $n$-th sample, where $\hat{y}^{n}_{i} = 0, i\in\{0,1,...,C\}$ except $\hat{y}^{n}_{c} = 1$.

To perform hard sample mining in a batch, we adopt triplet loss~\cite{triplet} denoted as $\mathcal{L}_\mathit{tri}\: (f)$, where $f$ indicates the input feature:

\begin{equation}
        \label{eq:trplt}
        \begin{split}
            \mathcal{L}_\mathit{tri}\: (f) =\: \: \, \sum_{i=1}^{P} \sum_{a=1}^{K}[m+ \overbrace{\mathop{\max}_{p=1,...,K}D(f_{a}^{i}, f_{p}^{i})}^{\mathit{hardest}\: \mathit{positive}} \\
            -\overbrace{\mathop{\min}_{n=1,...,K}D(f_{a}^{i},f_{n}^{i})}^{\mathit{hardest}\: \mathit{negative}}].
        \end{split}
\end{equation}

We use a combination of both of these losses to constrain and optimize the feature extraction capability of the overall network, defined as $\mathcal{L}_{ReID}$, 
\begin{equation}
  \mathcal{L}_{ReID} = \mathcal{L}_{id} + \mathcal{L}_{tri}.
\end{equation}
Finally, the total loss in our framework can be defined as,
\begin{equation}
\begin{split}
\mathcal{L}_{total} = (1 - \lambda) \mathcal{L}_{ReID}^{ViT} + \lambda \mathcal{L}_{ReID}^{CEM} + \mathcal{L}_{ReID}^{P}.
\end{split}
\label{eq:total_loss}
\end{equation}

\section{Experiments}

\subsection{Dataset and Evaluation Protocols}
To evaluate the effectiveness of the proposed CoEN on existing multi-spectral vehicle ReID public datasets, we utilize three multi-spectral vehicle ReID benchmarks and present comprehensive experimental results in this section. Specifically, MWVeID863~\cite{facenet} is a large-scale low-quality multi-spectral vehicle ReID dataset with intense flare interference. RGBNT100~\cite{li2020multi} is a large-scale and comprehensive multi-spectral vehicle ReID dataset. MSVR310~\cite{zheng2023cross} is a small-scale multi-spectral vehicle ReID dataset with complex scenarios. 

To ensure fair experimental evaluation, we follow previous research methods and use the mean Average Precision (mAP) and Cumulative Matching Characteristic (CMC), commonly used in ReID tasks, to evaluate our approach. CMC scores reflect retrieval precision, and we report Rank-1, Rank-5, and Rank-10 scores in our experiments. mAP measures the average precision of all queries, represented by the area under the precision-recall curve, providing a comprehensive assessment of recall and precision performance.

\subsection{Implementation Details}
Our model is implemented with the PyTorch toolbox.
We conduct experiments on one NVIDIA RTX 4090 GPU.
We employ pre-trained weights DeiT-B/16~\cite{deit2021} from the ImageNet~\cite{deng2009imagenet} as the parameters for our backbone ViT~\cite{dosovitskiy2020image}.
For data processing, the images were resized to $128$ $\times$ 256 for network input, with data augmentation methods including random cropping, horizontal flipping and random erasure.
We randomly select $4$ identities, each providing $8$ samples ($24$ images) as training samples for each training mini-batch.
To optimize our model, we use Stochastic Gradient Descent (SGD) optimizer to optimize the network with the initial learning rate as $3\times10^{-3}$ for WMVeID863~\cite{facenet} and $1\times10^{-3}$ for RGBNT100~\cite{li2020multi} and MSVR310~\cite{zheng2023cross}, both with a momentum of $0.9$ and a weight decays of $1\times10^{-4}$ at total $120$ epochs.
In evaluation, we concatenate the features extracted from three parallel branches as the final representation for a sample, for the feature matching. In the CEM, $\gamma$ is set to $0.5$. For the ReID loss, $\lambda$ is set to $0.5$.

\begin{table*}[!ht]
\caption{Comparison with state-of-the-art methods for multi-spectral ReID on RGBNT100 and MSVR310 (in \%). Both single-spectral and multi-spectral methods are included.}
\centering
\resizebox{0.8\linewidth}{!}{
\begin{tabular}{cccccccc}
\hline
\multirow{2}{*}{} & \multirow{2}{*}{Methods}& \multirow{2}{*}{Ref} & \multirow{2}{*}{Bac} & \multicolumn{2}{c}{RGBNT100}  & \multicolumn{2}{c}{MSVR310}\\ \cline{5-8} 
    & & &           & mAP & R-1 & mAP & R-1 \\ \hline
\multirow{8}{*}{Single}  
    &PCB~\cite{sun2018beyond}           &ECCV'18  & CNN & 57.2 & 83.5 & 23.2 & 42.9 \\
    &MGN~\cite{wang2018learning}        &ACMMM'18 & CNN & 58.1 & 83.1 & 26.2 & 44.3 \\
    &DMML~\cite{chen2019deep}           &ICCV'18  & CNN & 58.5 & 82.0 & 19.1 & 31.1 \\
    &BoT~\cite{luo2019bag}              &CVPR'19  & CNN & 78.0 & 95.1 & 23.5 & 38.4 \\
    &OSNet~\cite{zhou2019omni}          &ICCV'19  & CNN & 75.0 & 95.6 & 28.7 & 44.8 \\
    &Circle Loss~\cite{sun2020circle}   &CVPR'20  & CNN & 59.4 & 81.7 & 22.7 & 34.2 \\
    &HRCN~\cite{zhao2021heterogeneous}  &ICCV'21  & CNN & 67.1 & 91.8 & 23.4 & 44.2 \\
    &AGW~\cite{ye2021deep}              &TPAMI'21 & CNN & 73.1 & 92.7 & 28.9 & 46.9 \\
    &TransReID~\cite{he2021transreid}   &ICCV'21  & ViT & 75.6 & 92.9 & 18.4 & 29.6 \\ \hline
\multirow{8}{*}{Multi}
    &HAMNet~\cite{li2020multi}          &AAAI'20  & CNN & 74.5 & 93.3 & 27.1 & 42.3 \\
    &PFNet~\cite{zheng2021robust}       &AAAI'21  & CNN & 68.1 & 94.1 & 23.5 & 37.4 \\
    &IEEE~\cite{wang2022interact}       &AAAI'22  & CNN & 61.3 & 87.8 & 21.0 & 41.0 \\
    &CCNet~\cite{zheng2023cross}        &INFFUS'23& CNN & 77.2 & 96.3 & 36.4 & \underline{55.2} \\
    &TOP-ReID~\cite{topreid}         &AAAI'24  & ViT & 81.2 & 96.4 & 35.9 & 44.6 \\
    &EDITOR~\cite{magictoken}        &CVPR'24  & ViT & 82.1 & 96.4 & 39.0 & 49.3 \\
    &FACENet~\cite{facenet}          &INFFUS'25& ViT & 81.5 & \underline{96.9} & 36.2 & 54.1 \\
    &DeMo~\cite{wang2025DeMo}       &AAAI'25  & ViT & \underline{82.4} & 96.0 & \underline{39.1} & 48.6  \\ 
    &\bf{CoEN}                       &Ours     & ViT & \bf{84.6} & \bf{98.5} & \bf{44.0} & \bf{63.1}  \\ \hline
\multirow{4}{*}{Multiple}
    &DeMo~\cite{wang2025DeMo}       &AAAI'25  & CLIP& 86.2 & \bf{97.6} & \underline{49.2} & 59.8  \\
    &MambaPro~\cite{wang2025MambaPro}   &AAAI'25  & CLIP& 83.9 & 94.7 & 47.0 & 56.5   \\
    & IDEA~\cite{wang2025idea}      &CVPR'25  & CLIP  &\bf{87.2} & 96.5& 47.0 & \underline{62.4}   \\
    &\bf{CoEN}                       &Ours     & CLIP &\underline{86.4} & \underline{96.6} & \bf{52.2} & \bf{69.5}  \\
    \hline
\end{tabular}
}
\label{Tab:Sotaof100and310}
\end{table*}
\subsection{Comparison with State-of-the-Art Methods}
We evaluate the performance of our proposed method 
against various state-of-the-art ReID methods on the low-quality multi-spectral vehicle ReID dataset WMVeID863 \cite{facenet} and traditional multi-spectral vehicle ReID datasets RGBNT100~\cite{li2020multi} and MSVR310 \cite{zheng2023cross}, respectively.
For single-spectral methods, we simply extend them to multi-spectral versions for comparison by expanding the single-branch network into multiple branches and adding a corresponding loss function to each branch. 
\\
\textbf{Comparison on WMVeID863.}
As shown in Table~\ref{Tab:Sotaof863}, we evaluate the performance of our proposed CoEN method with both single-spectral and multi-spectral methods for multi-spectral vehicle ReID on WMVeID863~\cite{facenet}.
In contrast to multi-spectral methods, most current single-spectral methods do not fully utilize the complementary information across multi-spectral, even though they perform well on individual spectra.
It is noteworthy that Transformer-based methods significantly outperform CNN-based approaches due to their superior global information extraction capabilities.
Among these methods, TransReID~\cite{he2021transreid} stands out with a 67.0\% mAP.
For Transformer-based multi-spectral methods, the traditional multi-spectral enhancement methods TOP-ReID~\cite{topreid}  and DeMo~\cite{wang2025DeMo} achieve a notable improvement.
Notably, FACENet~\cite{facenet} achieves a superior performance of 69.8\% by using the flare-immune TIR as the primary spectrum to enhance the remaining flare-affected spectra with corrupted information.
However, our CoEN overcomes the problem of information loss in multiple spectra in different scenes and outperforms FACENet~\cite{facenet} by 1.6\% in mAP and 2.2\% in Rank-1.
Specifically, our CoEN achieves optimal results even when compared to methods~\cite{wang2025DeMo,wang2025MambaPro} using CLIP as a backbone.
%
This demonstrates the superiority of our approach in addressing the challenges of low-quality multi-spectral vehicle ReID tasks.
\\
\textbf{Comparison on RGBNT100 and MSVR310.}
As shown in Table~\ref{Tab:Sotaof100and310}, although some single-spectral methods have achieved good performance, such as BoT~\cite{luo2019bag} and AGW~\cite{ye2021deep} on RGBNT100~\cite{li2020multi} and MSVR310~\cite{zheng2023cross}, respectively, the single-spectral methods are also slightly insufficient when compared with the multi-spectral methods. Among current multi-spectral methods, Transformer-based EDITOR~\cite{magictoken} achieves 82.1\% mAP on the RGBNT100~\cite{li2020multi} dataset and 39.0\% mAP on the MSVR310~\cite{zheng2023cross} dataset. DeMo~\cite{wang2025DeMo} achieves the best performance by decoupling different spectra into non-overlapping forms. However, our Transformer-based CoEN surpasses DeMo~\cite{wang2025DeMo}, achieving a 2.2\% and 4.9\% improvement in mAP accuracy for the RGBNT100~\cite{li2020multi} and MSVR310~\cite{zheng2023cross} datasets, respectively. 
Furthermore, when we replace the CoEN backbone network with CLIP, it shows superior performance in the MSVR310~\cite{zheng2023cross} datasets. Meanwhile, despite the lack of corresponding textual annotations, our CoEN is still capable of supplementing discriminative information for individual spectra and achieves competitive performance on the RGBNT100~\cite{li2020multi} dataset.

These results validate the superiority of the CoEN approach in complex scenarios and on large-scale multi-spectral vehicle datasets, demonstrating its effectiveness over existing solutions.

\subsection{Ablation Studies}

\begin{table}[t]
\caption{Performance comparison with different components. $Pri_{E}$ and $Pro_{E}$ stand for Primary-based Enhancement Strategy and Proxy-based Enhancement Strategy respectively}
\centering
\setlength{\tabcolsep}{4pt}

\resizebox{\linewidth}{!}{

\begin{tabular}{c|cccc|cccc}
\hline
\multirow{2}{*}{} & \multirow{2}{*}{PG} & \multirow{2}{*}{DQSM} & \multicolumn{2}{c|}{CEM} & \multicolumn{4}{c}{WMVeID863} \\
  &           &           & $Pri_{E}$   & $Pro_{E}$   & mAP  & R-1  & R-5  & R-10 \\ \hline
(a) & \ding{53} & \ding{53} & \ding{53} & \ding{53} & 66.6 & 74.0 & 80.6 & 82.1 \\
(b) & \ding{51} & \ding{53} & \ding{53} & \ding{53} & 68.6 & 75.6 & 81.0 & 82.9 \\
(c) & \ding{51} & \ding{51} & \ding{53} & \ding{51} & 70.4 & 77.4 & 83.1 & 85.4 \\
(d) & \ding{51} & \ding{51} & \ding{51} & \ding{53} & 70.3 & 78.3 & 82.4 & 85.4 \\
(e) & \ding{51} & \ding{51} & \ding{51} & \ding{51} & \bf{71.4} & \bf{79.2} & \bf{83.5} & \bf{85.6} \\ \hline
\end{tabular}
}
\label{Tab:ablation}
\vspace{8pt}
\end{table}

We conduct ablation studies on the WMVeID863~\cite{facenet} dataset on the ViT-based backbone to validate the proposed components.
\\
\textbf{Effect of Key Components.}
Table~\ref{Tab:ablation} presents a performance comparison across different model configurations. Model (a) serves as the ViT~\cite{dosovitskiy2020image} baseline.
Model (b) incorporates PG, leading to a 1.7\% increase in mAP, indicating that the PG module effectively aggregates the residual vehicle key cues from the multi-spectrum, generating spectrally fused features rich in identification information.
Furthermore, Model (c) introduces DQSM and primary-based enhancement, achieving additional performance gains by accurately filtering high-quality spectra from various scenes through correlation analysis and enhancing low-quality spectra with missing detail information.
Model (d) solely employs the proxy-based enhancement strategy to enhance and compensate for all spectral vehicle cues, resulting in improved performance.
By integrating all components of Model (e) and applying the collaborative enhancement strategy, our CoEN achieves superior performance, demonstrating the effectiveness of the collaborative enhancement strategy in low-mass multi-spectral vehicle ReID.

\begin{table}[h]
\caption{Effect of different proxy features generation methods in PG.}
\centering
\resizebox{\linewidth}{!}{
\begin{tabular}{ccccccc}
\hline
\multirow{2}{*}{Generation Methods} & \multicolumn{4}{c}{WMVeID863} \\ \cline{2-5} 
                  & mAP  & R-1    & R-5    & R-10   \\ \hline
              Sum & 70.7 & 78.5   & 82.0   & 83.8    \\
         Direct Proj & 70.8 & 78.1   & 82.7   & 84.5    \\
\textbf{Progressive Proj} & \bf{71.4} & \bf{79.2}   & \bf{83.5}   & \bf{85.6}    \\ \hline
\end{tabular}
}
\label{tab:PG_generation}
\end{table}

\begin{table}[h]
\caption{Effect of proxy features in PG. All spectral inputs during inference.}
\centering
\resizebox{\linewidth}{!}{
\begin{tabular}{ccccccc}
\hline
\multirow{2}{*}{Inference Methods} & \multicolumn{4}{c}{WMVeID863} \\ \cline{2-5} 
                            & mAP  & R-1    & R-5    & R-10   \\ \hline
                        RGB & 57.8 & 62.5   & 74.9   & 80.2    \\
                        NIR & 58.6 & 64.4   & 77.0   & 82.0    \\
                        TIR & 66.1 & 71.5   & 78.1   & 81.3    \\
                      Proxy & 67.9 & 74.6   & 79.9   & 82.9    \\
                RGB-NIR-TIR & 69.7 & 77.4   & 82.7   & 84.9    \\
      \textbf{RGB-NIR-TIR-Proxy}& \bf{71.4} & \bf{79.2}   & \bf{83.5}   & \bf{85.6}    \\  \hline
\end{tabular}
}
\label{tab:PG_ablation}
\end{table}

\noindent\textbf{Effect of Proxy Generator.}
As shown in Table~\ref{tab:PG_generation}, to validate further the effectiveness of the progressive dimensionality reduction strategy in the PG module, we compare various spectral fusion feature generation methods, including element addition and direct dimensionality reduction. The results indicate that these methods fail to fully integrate the remaining key identity information in the multi-spectrum, resulting in information confusion and negatively impacting the generation process. In contrast, the progressive projection strategy successfully maps multi-spectral information into a unified identity space through hierarchical spatial mapping, significantly enhancing model performance.

As shown in Table~\ref{tab:PG_ablation}, we further validate the effectiveness of the proxy generated by the PG module. During training, all spectra are used for training, while during testing, all spectra are input into the model to extract enhanced multi-spectral features from the test set using the generated spectral fusion features. Inference is then performed separately on all extracted spectral features. The results demonstrate that the generated proxy contain more vehicle discrimination information than individual spectra. Furthermore, combining all features for inference further improves the performance of our method.
\begin{table}[h]
\caption{Effect of different primary spectral selection methods in DQSM.}
\centering
\resizebox{\linewidth}{!}{
\begin{tabular}{ccccccc}
\hline
\multirow{2}{*}{Methods}  & \multicolumn{4}{c}{WMVeID863} \\ \cline{2-5} 
                          & mAP  & R-1    & R-5    & R-10   \\ \hline
                 w/o DQSM & 70.4 & 78.3   & 82.4   & 85.4   \\
RGB$\rightarrow$(NIR,TIR) & 69.9 & 76.9   & 82.4   & 84.7    \\
NIR$\rightarrow$(RGB,TIR) & 70.0 & 77.0   & 82.2   & 84.2    \\
TIR$\rightarrow$(RGB,NIR) & 70.6 & 77.8   & 83.1   & 84.9    \\ 
         \textbf{w/ DQSM} & \bf{71.4} & \bf{79.2}   & \bf{83.5}  & \bf{85.6}    \\\hline
\end{tabular}
}
\label{tab:DQSM_ablation}
\end{table}
\\
\textbf{Effect of Dynamic Quality Sort Module.}
As shown in Table~\ref{tab:DQSM_ablation}, we evaluate the effectiveness of DQSM. We first validate the experimental results when the primary spectrum selected by DQSM is not used for enhancement and compare them with scenarios where a specific spectrum is fixed for enhancement and where high-quality spectra are dynamically selected for enhancement. The results indicate that fixing a primary spectrum to enhance other spectra has significant limitations and can even introduce noise during feature fusion, degrading model performance. In contrast, DQSM dynamically selects high-quality spectra by analyzing the remaining vehicle information content, effectively supplementing low-quality spectra with missing identity cues and improving overall performance.
\begin{table}[t]
\caption{Effects of collaborative enhancement strategy in CEM.}
\centering
\setlength{\tabcolsep}{4pt}

\resizebox{1.0\linewidth}{!}{
\begin{tabular}{ccccccc}
\hline
\multirow{2}{*}{Methods} & \multicolumn{4}{c}{WMVeID863} \\ \cline{2-5} 
                          & mAP  & R-1    & R-5    & R-10   \\ \hline
                  w/o CEM & 68.6 & 75.6   & 81.0   & 82.9   \\
    w/ TPM~\cite{topreid} & 69.8 & 77.6   & 82.2   & 83.8   \\ 
          \textbf{w/ CEM} & \bf{71.4} & \bf{79.2}   & \bf{83.5}   & \bf{85.6}   \\\hline
\end{tabular}
}
\label{tab:CEM_ablation}
\end{table}
\\
\textbf{Effect of Collaborative Enhancement Module.}
As shown in Table~\ref{tab:CEM_ablation}, we conduct comparative experiments between CEM and other multi-spectral enhancement methods. The results demonstrate that CEM effectively supplements and enhances missing vehicle discrimination information in spectra of varying qualities by combining primary and proxy-based enhancement strategies, thus significantly improving model performance.

\subsection{Hyperparameter Analysis}
Our approach involves two key hyperparameters: $\gamma$ in Eq.~\eqref{eq:pe},~\eqref{eq:fe} and $\lambda$ in Eq.~\eqref{eq:total_loss}. The hyperparameter $\gamma$ represents the dropout rate in the attention mechanism, which regulates the utilization of enhanced features in the CEM and significantly influences the generalization ability of the model. The hyperparameter $\lambda$ is the weight of the loss function, responsible for balancing the ReID loss before and after enhancement and managing the prioritization of ViT~\cite{dosovitskiy2020image} and CEM output features during parameter optimization.

\begin{figure}
\centering
\includegraphics[width=1\linewidth]{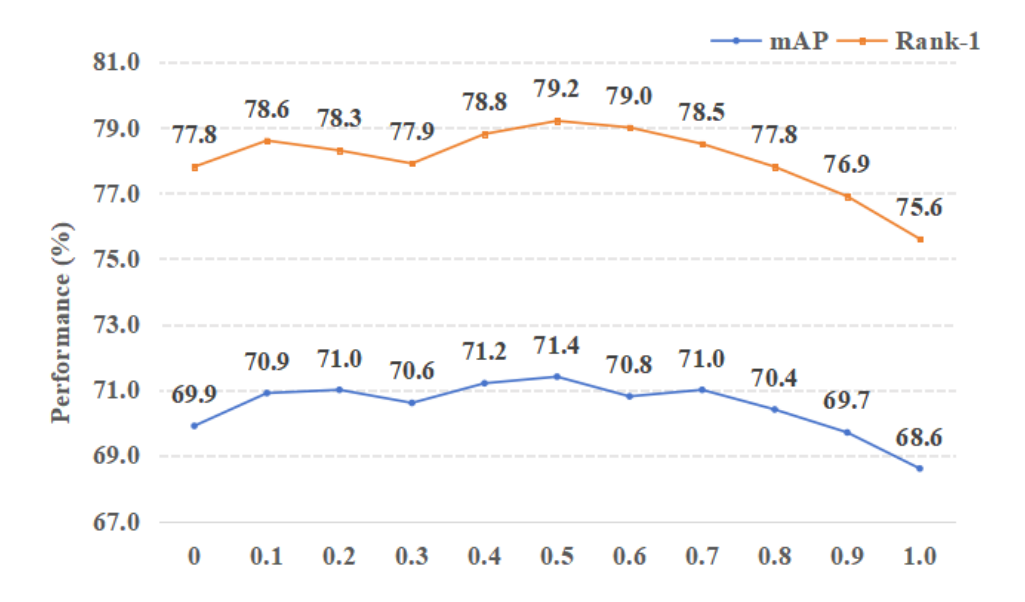}
  \vspace{-1mm}
  \caption{Effect of $\gamma$ on mAP (\%) and Rank-1 (\%) on WMVeID863}
  \label{fig:r_weight}
  \vspace{-4mm}
\end{figure}

\begin{figure}
\centering
\includegraphics[width=0.95\linewidth]{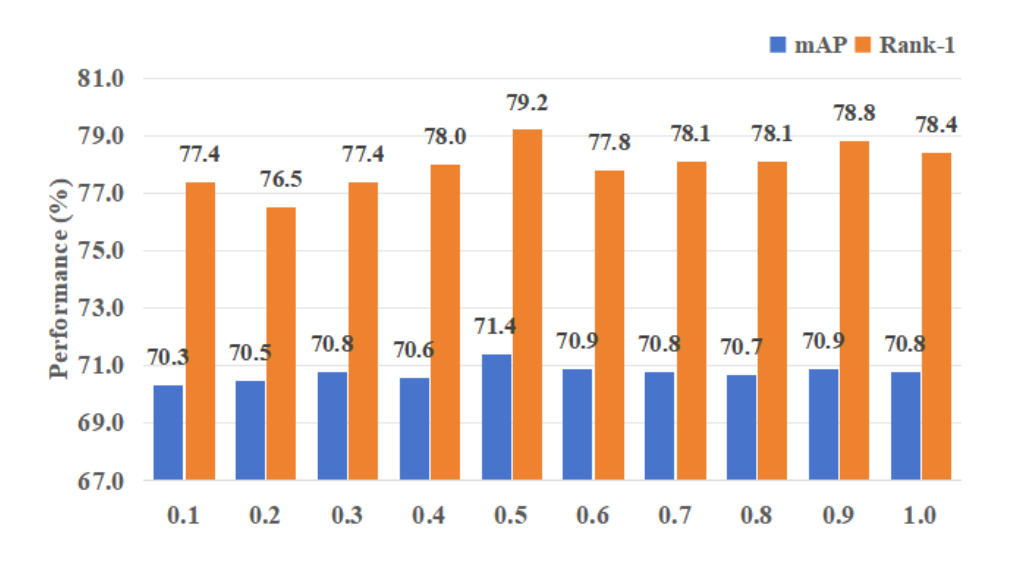}
  \vspace{-1mm}
  \caption{Effect of $\lambda$ on mAP (\%) and Rank-1 (\%) on WMVeID863}
  \label{fig:reid_param}
  \vspace{-4mm}
\end{figure}

To examine the impact of hyperparameters $\gamma$ and $\lambda$ on model performance, we conduct systematic quantitative experiments, tuning their values within the range of $0.1$ to $1.0$. As shown in Fig.~\ref{fig:r_weight}, increasing the value of $\gamma$ helps the model learn more generalized features. However, excessively high $\gamma$ values result in the loss of discriminative features, while too low values cause overfitting and limit generalization ability. Fig.~\ref{fig:reid_param} illustrates the impact of $\lambda$: excessively high or low values can cause the model to focus improperly on features, disrupting the equilibrium of the optimization process. The experimental results indicate that the model achieves optimal performance when both $\lambda$ and $\gamma$ are set to $0.5$.

\begin{figure}
\centering
\includegraphics[width=1\linewidth]{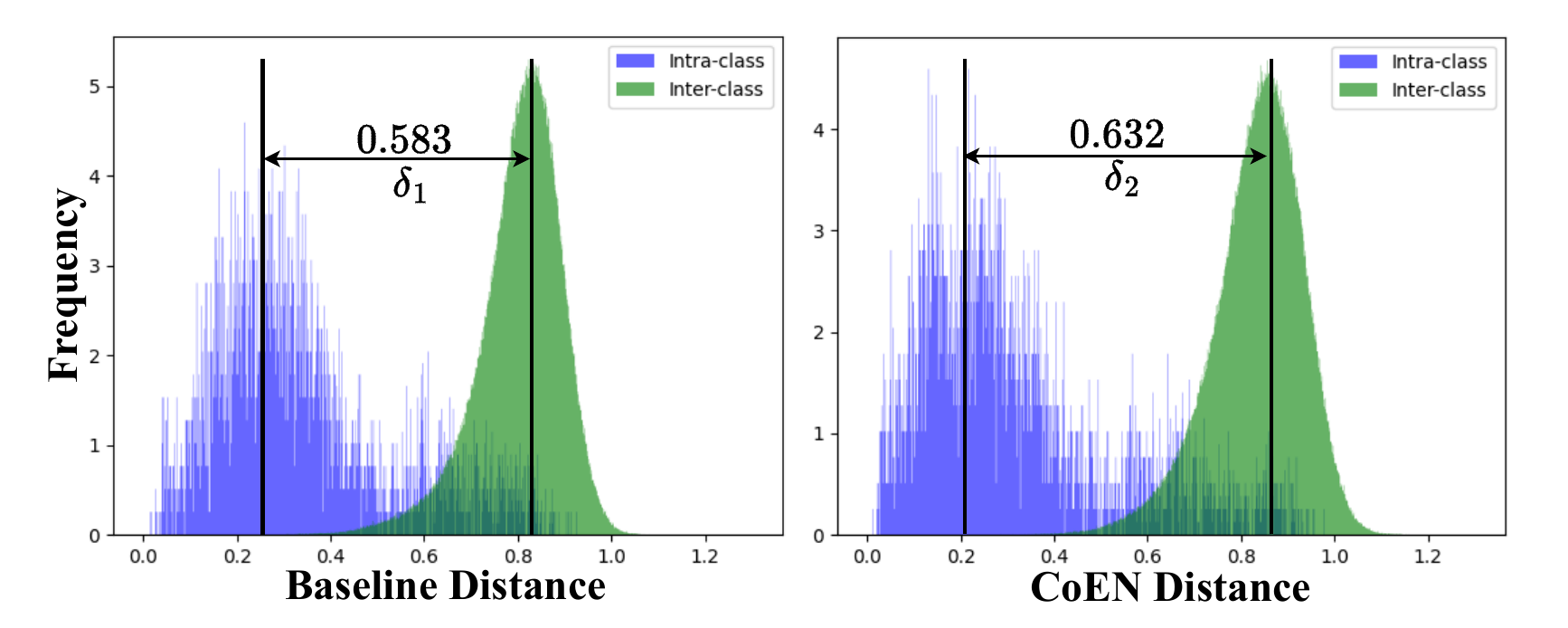}
  \caption{The distributions of the two types of distances between the multi-spectral vehicle features on WMVeID863.}
  \label{fig:distance}
\end{figure}

\begin{figure}
\centering
\includegraphics[width=1\linewidth]{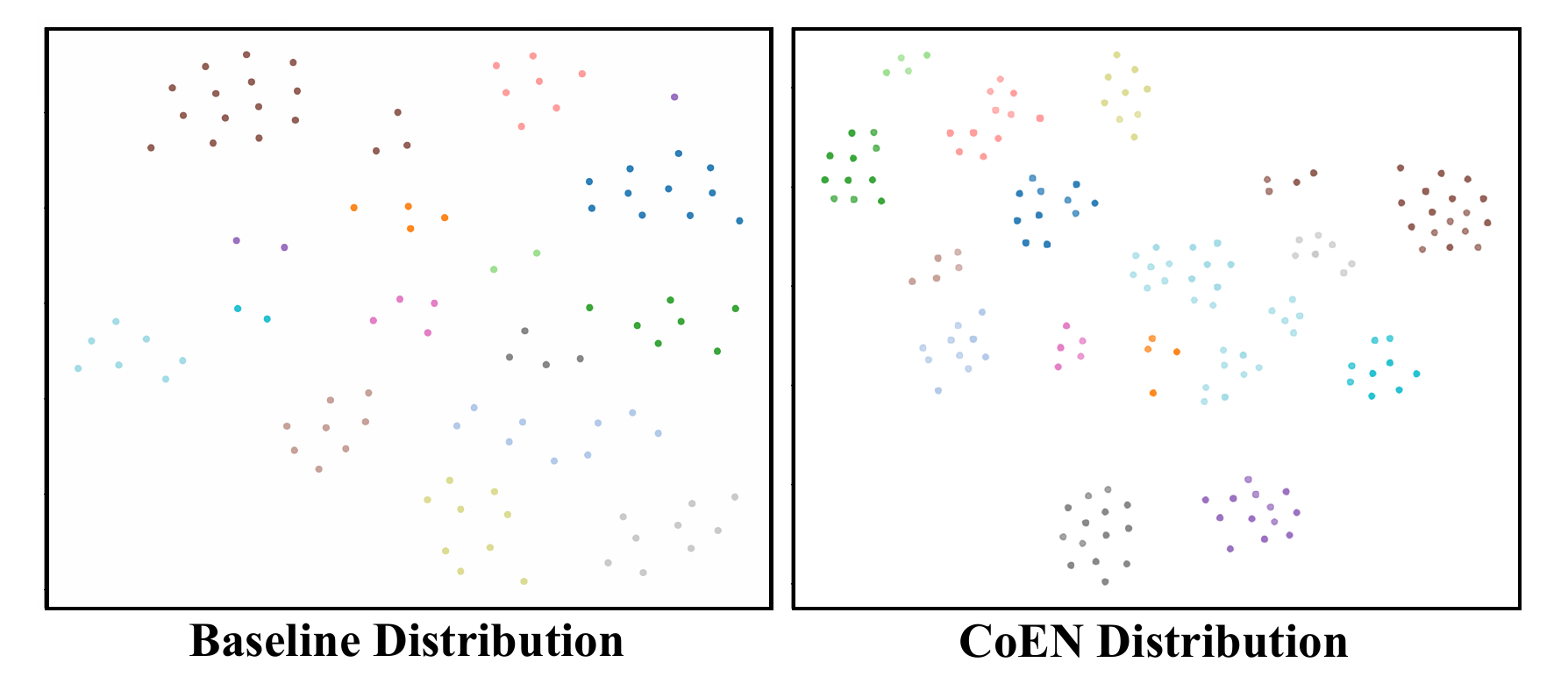}
  \caption{Comparison of feature distributions with T-SNE~\cite{van2008visualizing} on WMVeID863. Different colors represent different identities.}
  \label{fig:tsne}
\end{figure}

\begin{figure}
\centering
\includegraphics[width=1\linewidth]{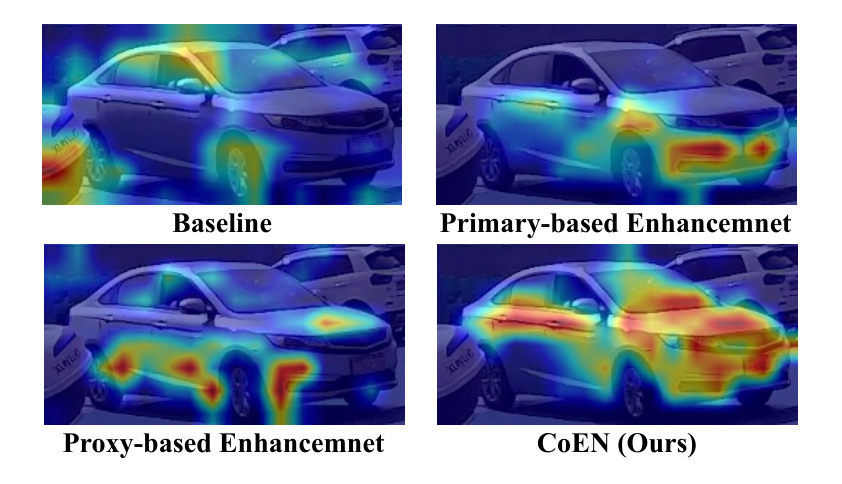}
  \vspace{-1mm}
  \caption{Class activation map visualization on WMVeID863 for different enhancement strategies.}
  \label{fig:gradcam_CEM}
  \vspace{-4mm}
\end{figure}

\subsection{Visualization Analysis}

\noindent\textbf{Distance Distributions.}
To validate the effectiveness of CoEN, we perform frequent statistical analyses of inter-identity and intra-identity distances. As shown in Fig.~\ref{fig:distance}, we compare the distance distributions between the baseline method and the proposed method. The analysis reveals that the proposed method demonstrates significant improvements in separating inter-identity and intra-identity distances, effectively enhancing the discriminative ability of features.

\noindent\textbf{Feature Distributions.}
We employ the T-SNE~\cite{van2008visualizing} dimensionality reduction technique to visualize and analyze the feature distributions of $15$ vehicles. As shown in Fig.~\ref{fig:tsne}, compared to the baseline method, CoEN significantly reduces feature variations within the same identity, resulting in more compact feature distributions and clearer inter-class separation. These results indicate that CoEN not only enhances the discriminative properties of features but also improves their robustness.

\begin{figure}
\centering
\includegraphics[width=\columnwidth]{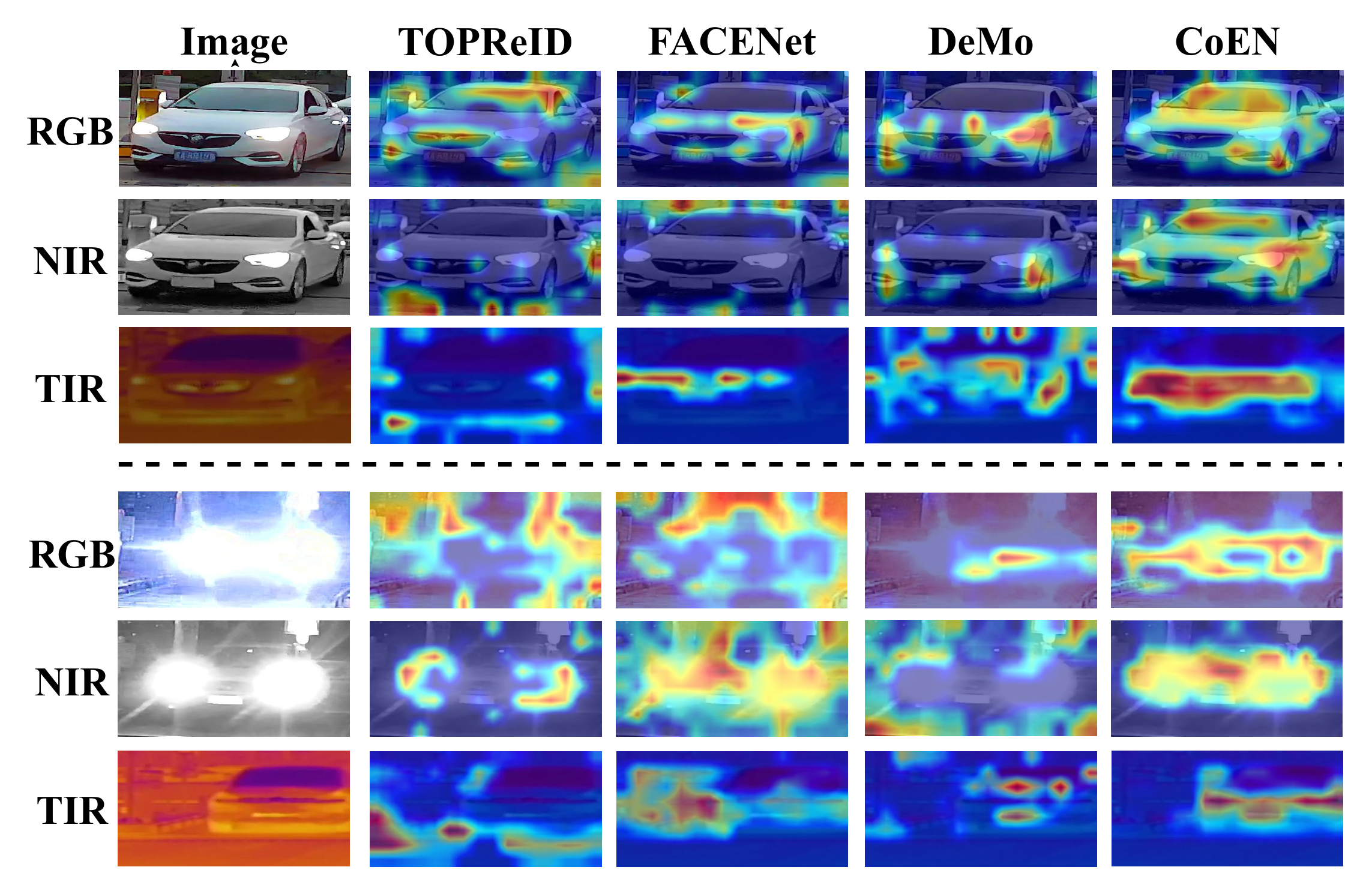}
  \vspace{-1mm}
  \caption{Class activation map visualization on WMVeID863 comparing different methods.}
  \label{fig:gradcam}
  \vspace{-4mm}
\end{figure}

\noindent\textbf{Class Activation Map Visualization.}
To validate the effectiveness of CoEN in addressing spectrally missing information for low-quality multi-spectral vehicle ReID, we visualize the results before and after applying the proposed method using Grad-CAM~\cite{gradcam}. 
As shown in Fig.~\ref{fig:gradcam_CEM}, we visualize the class activation maps for three scenarios: the primary-based enhancement, the proxy-based enhancement, and the proposed collaborative enhancement strategy. The results demonstrate that while both enhancement methods can partially supplement vehicle identity information in the spectra, the proposed CoEN effectively collaborates the two strategies, introducing richer discriminative cues into the spectra. 
As shown in Fig.~\ref{fig:gradcam}, we compare the visualization results of CoEN with existing multi-spectral vehicle ReID methods. Compared to the traditional method TOP-ReID~\cite{topreid} and Demo~\cite{wang2025DeMo}, the results indicate that CoEN avoids introducing low-quality spectral noise and extracts vehicle identity information more accurately. Additionally, compared to FACENet~\cite{facenet}, CoEN not only addresses spectral low-quality issues in flare scenes but also dynamically adapts to other scenarios.
\section{Conclusion}
In this work, we propose a Collaborative Enhancement Network (CoEN) to address the low-quality multi-spectral vehicle ReID problem. The network first employs a Proxy Generator to generate a fused proxy representation with spatial consistency by aggregating the key information from multiple spectra. The Dynamic Quality Sort Module then evaluates individual spectra based on the information relevance, assigns quality scores, and identifies the primary spectra. Finally, the Collaborative Enhancement Module combines primary-based enhancement and proxy-based enhancement strategies to effectively recover missing information in spectra of different quality, thereby improving the quality of the multi-spectral feature representation. Extensive experiments on three multi-spectral vehicle benchmark datasets demonstrate the effectiveness of our approach. In the future, we will explore incorporating modality-independent textual information to more precisely localize discriminative key features in multi-spectral data and enhance the utilizing of critical cues.

\section*{Data Availability Statement}
The data used to support the findings of this study are included in the paper.
\section*{Declaration of Competing Interest}
No potential conflict of interest was reported by the authors.

\section*{Author Statement}
Aihua Zheng: Conceptualization of this study and Methodology and Investigation. 
Yongqi Sun: Writing-Original Draft and Validation and Visualization.
Zi Wang: Writing Review and Editing. 
Chenglong Li: Formal Analysis and Data Curation. 
Jin Tang: Resources and interpretation of data.

\section*{Acknowledgements}
This research is partly supported by the National Natural Science Foundation of China
(No. 62372003), the Natural Science Foundation of Anhui Province (No.2308085Y40 and No. 2208085J18) and the University Synergy Innovation Program of Anhui Province (No. GXXT-2022-036).

\bibliographystyle{cas-model2-names}
\bibliography{ref}
\bio{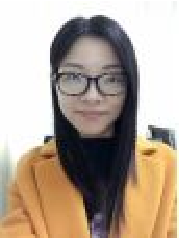}
{Aihua Zheng} received B.Eng. degrees and finished Master-Docter combined program in Computer Science and Technology from Anhui University of China in 2006 and 2008, respectively. And received Ph.D. degree in computer science from University of Greenwich of UK in 2012. She visited University of Stirling and Texas State University during June to September in 2013 and during September 2019 to August 2020 respectively. She is currently an Professor and PhD supervisor at the School of Artificial Intelligence, Anhui University.
Her main research interests include vision based artificial intelligence and pattern recognition. Especially on person/vehicle re-identification, audio visual computing, and multi-modal intelligence.
\endbio

\bio{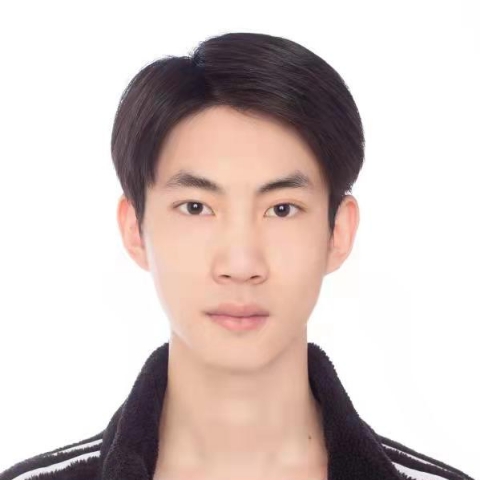}
{Yongqi Sun} 
    	received his B.Eng. degree in 2022 and is currently pursuing the M.Eng degree in the School of Artificial Intelligence, Anhui University, Hefei, China. His research interests include Computer Vision, Multi-modal Intelligence and Vehicle Re-identification.
\endbio

\bio{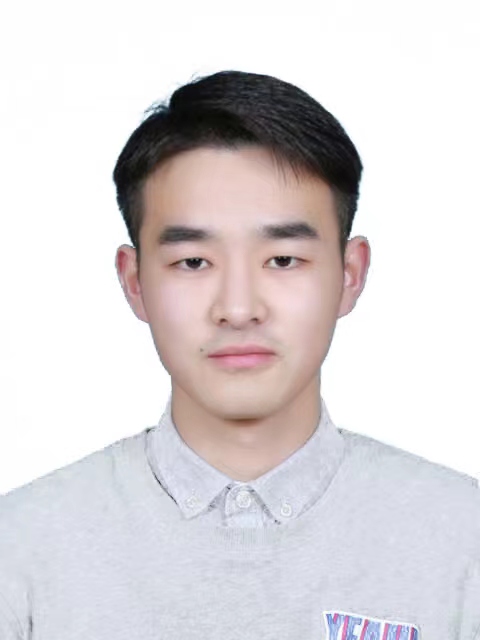}
{Zi Wang}
	received his B.Eng degrees and completed the Master-Doctor combined program in the School of Computer Science and Technology, Anhui University, Hefei, China. He works at the School of Biomedical Engineering, Anhui Medical University, Hefei, China. He is primarily engaged in research on computer vision, medical image processing, and multi-modal learning.
\endbio

\bio{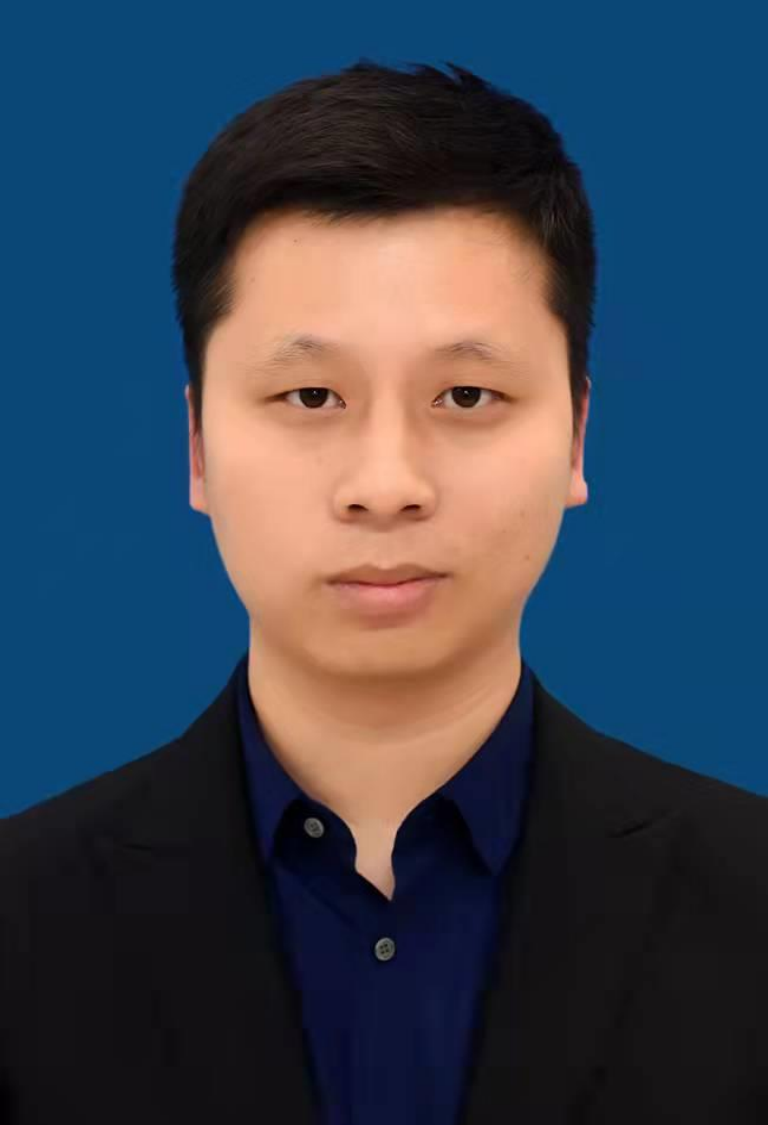}
{Chenglong Li}
		received the M.S. and Ph.D. degrees from the School of Computer Science and Technology, Anhui University, Hefei, China, in 2013 and 2016, respectively. From 2014 to 2015, he worked as a Visiting Student with the School of Data and Computer Science, Sun Yat-sen University, Guangzhou, China. He was a postdoctoral research fellow at the Center for Research on Intelligent Perception and Computing (CRIPAC), National Laboratory of Pattern Recognition (NLPR), Institute of Automation, Chinese Academy of Sciences (CASIA), China. 
		He is currently an Professor and PhD supervisor at the School of Artificial Intelligence, Anhui University. His research interests include computer vision and deep learning. He was a recipient of the ACM Hefei Doctoral Dissertation Award in 2016.
\endbio

\bio{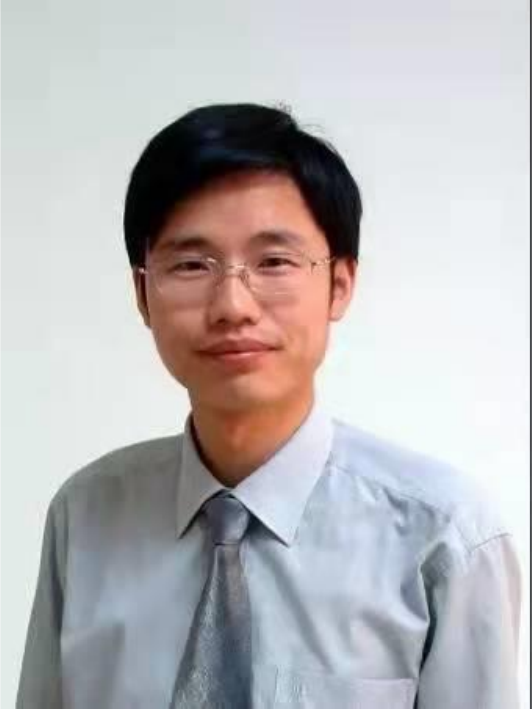}
{Jin Tang}
		received the B.Eng. degree in automation and the Ph.D. degree in Computer Science from Anhui University, Hefei, China, in 1999 and 2007, respectively. He is currently a Professor and PhD supervisor with the School of Computer Science and Technology, Anhui University. 
		His research interests include computer vision, pattern recognition, and machine learning.
\endbio

\end{document}